\documentclass[11pt]{article}

\PassOptionsToPackage{sort}{natbib}
\usepackage[final]{acl}

\usepackage{times}
\usepackage{latexsym}
\usepackage[T1]{fontenc}
\usepackage[utf8]{inputenc}
\usepackage{textcomp}   
\usepackage{microtype}
\usepackage{inconsolata}

\usepackage{amsmath}
\usepackage{amssymb}

\usepackage{longtable}
\usepackage{booktabs}
\usepackage{array}     
\usepackage{calc}      
\usepackage{multirow}  
\usepackage{float}     
\usepackage{pdflscape} 

\usepackage{etoolbox}
\newif\ifWasInTwoColumn
\makeatletter
\BeforeBeginEnvironment{longtable}{%
  \if@twocolumn\WasInTwoColumntrue\onecolumn\else\WasInTwoColumnfalse\fi}
\AfterEndEnvironment{longtable}{%
  \ifWasInTwoColumn\twocolumn\fi}
\makeatother

\def\fnum@table{\tablename~\thetable}

\usepackage{graphicx}
\makeatletter
\def\maxwidth{\ifdim\Gin@nat@width>\linewidth\linewidth\else\Gin@nat@width\fi}
\def\maxheight{\ifdim\Gin@nat@height>\textheight\textheight\else\Gin@nat@height\fi}
\def\fps@figure{htbp}
\makeatother

\IfFileExists{upquote.sty}{\usepackage{upquote}}{}  
\usepackage{fancyvrb}

\date{}

\makeatletter
\newsavebox\pandoc@box
\newcommand*\pandocbounded[1]{%
  \sbox\pandoc@box{#1}%
  \Gscale@div\@tempa{\linewidth}{\wd\pandoc@box}%
  \ifdim\@tempa\p@<\p@\scalebox{\@tempa}{\usebox\pandoc@box}\else#1\fi%
}
\makeatother

\providecommand{\tightlist}{%
  \setlength{\itemsep}{0pt}\setlength{\parskip}{0pt}}

\title{Exemplars in Disguise: Pure Exemplar Models Mimic
Abstraction-First Learning}

\author{
  Zachary Nicholas Houghton \\
      University of Oregon \\ Vail Systems, Inc. \\
        \texttt{znh@uoregon.edu}
     \And
  Vsevolod Kapatsinski \\
      University of Oregon \\
      }

\begin{document}
\maketitle

\begin{abstract}
Whether idiosyncratic, item-specific knowledge is learned before
abstract class-level generalizations, or vice versa, is a central
question in language learning, with exemplar and abstraction-based
theories making opposite predictions. Recent methods have claimed to
show that, at least for large language models, abstract knowledge is
learned first. We show that these methods fall short: pure memorizer
models with no abstract representations can appear, by the same
criteria, to learn either item-specific or class-level knowledge first,
depending on their sensitivity to individual observations, with the
transition point governed by the distributional properties of the input.
We further argue that the distinction between item-specific and abstract
knowledge may be ill-defined for distributed representations, as a
word's class-level properties may not be separable from its
item-specific properties.
\end{abstract}

\section{Introduction}\label{introduction}

Language learning draws on both abstract representations and
item-specific knowledge. A learner must learn the idiosyncrasies for
verbs like \emph{give} and \emph{run} and represent them as distinct
lexical items, while also discovering that \emph{give} and \emph{sell}
share class-level distributional preferences that set them apart from
motion verbs like \emph{go} and \emph{walk}. How these two levels of
knowledge interact, and crucially which emerges first during learning,
is a central question in cognitive science and linguistics.

A substantial body of work argues that abstract representations play a
central role in linguistic knowledge, enabling generalizations that
extend productively to novel items beyond what has been directly
observed
\citep{berkoChildsLearningEnglish1958, albright2003rulesvsanalogy, tomaselloConstructingLanguageUsagebased2005, goldberg2003constructionsnewtheoretical, kapatsinskiChangingMindsChanging2018, kapatsinskiWhatGrammarLike2014, kapatsinskiWhatStatisticsLearners2012, morgan2016abstractknowledgedirect, morgan2024productiveknowledgeitemspecific, houghton2025multiwordrepresentationsminds, milin2023learningperspective, harmon2017puttingoldtools}.
\citet{berkoChildsLearningEnglish1958} showed that children could
generate the plural of \emph{wug} (\emph{wugs}) without any prior
experience with the word, demonstrating that morphological knowledge
must be at least partly abstract and productive. In the domain of
phonology, \citet{albright2003rulesvsanalogy} showed that an
abstraction-based model out-performs other models in the learning of
English plural allomorphy. Indeed evidence for abstractions has been
pursued across every subfield in linguistics with each domain offering
distinct empirical windows into how abstract representations are
structured and learned
\citep{tomaselloConstructingLanguageUsagebased2005, goldberg2003constructionsnewtheoretical, kapatsinskiChangingMindsChanging2018, morgan2016abstractknowledgedirect, morgan2024productiveknowledgeitemspecific, houghton2025multiwordrepresentationsminds, houghton2025roleabstractrepresentations, pluth2026mechanisticinterpretabilityasr}.

Exemplar theories offer a competing account of how such knowledge is
represented, particularly in the categorization literature
\citep{nosofskyattentionsimilarity, nosofskyInvestigationsExemplarbasedConnectionist1992, ambridgeStoredAbstractionsRadical2020, bybee2001, bybee2003, stembergerFrequencyLexicalStorage1986, stembergerAreInflectedForms2004, bybee1999usagebasedphonology, bybee2002wordfrequencycontext, bybee2017mechanismschangegrammaticization, bybeeEffectUsageDegrees1999, pierrehumbert2001exemplardynamics, pierrehumbert2002wordspecificphonetics, mcmurray2002gradienteffects, racz2020morphologicalconvergence, goldrick2023exemplarmodels, gradoville2023futureexemplar}.
Rather than positing abstract representations, exemplar models argue
that categories are collections of stored instances, and that category
membership is determined by similarity to previously encountered
exemplars. This framework naturally accounts for the frequency effects
on gradient category membership: a robin is judged more bird-like than a
chicken not because it better fits an abstract representation, but
because it is more similar to the bulk of stored exemplars. Apparent
generalizations emerge from clusters of exemplars, with no explicit
abstraction required. In linguistics, usage-based approaches have
developed this intuition extensively \citep{bybee2001, bybee2003}: token
frequency strengthens the representation of specific forms, explaining
why high-frequency irregulars like \emph{went} and \emph{broke} resist
regularization, while type frequency of a pattern licenses analogical
extension to novel items without requiring stored abstractions.
\citet{stembergerFrequencyLexicalStorage1986} provided early
psycholinguistic evidence for frequency-sensitive storage \citep[see
also][]{stembergerAreInflectedForms2004}, showing that high-frequency
regular forms are retrieved holistically rather than generated
compositionally. \citet{ambridgeStoredAbstractionsRadical2020} extended
this reasoning to argument structure, arguing that even apparently
abstract grammatical patterns are best explained as emergent
generalizations over stored exemplars.

Between these extremes lie hybrid accounts in which abstract
representations exist but are themselves built from specific instances
\citep[e.g.,][]{ambridgeAbstractionsMadeExemplars2020, goldberg2003constructionsnewtheoretical, kapatsinskiChangingMindsChanging2018, stembergerFrequencyLexicalStorage1986, morgan2016abstractknowledgedirect, morganLevyModelingIdiosyncraticPreferences2015, morganLevyFrequencyDependentRegularization2016, liuMorganFrequencyDependentRegularization2020, houghton2025multiwordrepresentationsminds, houghton2026holisticstorageverbupphrases, houghtonMorganHolisticStorageCompoundNouns2023}.
The connectionist model of \citet{rumelhart1986learningtensesenglish}
was foundational here, and not simply because it reproduced the
behavioral data: the deeper significance was that its connection weights
jointly encoded both item-specific frequency knowledge and the
class-level regularity of the past-tense rule in the same distributed
representation, with no principled boundary between what is stored as an
exemplar and what is computed via abstract knowledge. This reframed the
debate: in a distributed representational system, class-level
abstractions and item-specific exemplar knowledge are not two separate
mechanisms but two levels of description of the same learned weight
structure. \citet{kapatsinskiWordsVersusRules2018} further developed
this perspective theoretically, arguing that frequency-sensitive
learning is inherently hybrid: because connection strengths are shaped
by the full history of individual exposures, the learned representation
simultaneously reflects item-specific token frequency and class-level
type frequency without requiring either to be computed separately;
\citet{kapatsinskiHierarchicalInferenceSound2021} provided a formal
instantiation of this in phonological learning, where hierarchical
inference over experience-tagged exemplars allocates credit
simultaneously to word-specific and phoneme-class-level representations.

The convergence on a hybrid account shifts the central question: rather
than asking whether abstract or exemplar-based representations exist,
the more pressing issue is which emerges first during learning. This
ordering question has direct consequences for theories of language
learning. If class-level abstract representations emerge first, early
generalization can scaffold subsequent item-specific learning. If
item-specific exemplars accumulate first, abstractions emerge bottom-up
through the gradual organization of individual instances.
\citet{tomaselloConstructingLanguageUsagebased2005} anticipated this
question at the developmental level, arguing that children's early
constructions are item-specific and that productive abstraction only
emerges gradually throughout learning. The two accounts thus make
opposite predictions about the temporal ordering of class-level versus
item-specific knowledge, predictions that are in principle detectable
from learning trajectories.

These questions have gained new empirical traction with the availability
of large language models, whose learning trajectories can be observed
directly over the course of training, making them a productive testing
ground for theoretical predictions about the ordering of class-level and
item-specific learning in distributed representations.

One such study exploring this was \citet{jian2026humanstransformerlms},
who investigated this directly by analyzing learning trajectories in
GPT-2 small over the course of pretraining. Their study tracked
next-token prediction distributions conditioned on argument structure
constructions across four verb classes: to-Dative verbs (e.g.,
\emph{give}, \emph{sell}), Motion verbs (e.g., \emph{go}, \emph{walk}),
Reciprocal verbs (e.g., \emph{speak}, \emph{meet}), and spray-load verbs
(e.g., \emph{spray}, \emph{load}). Using Jensen-Shannon divergence
(\(D_\text{JS}\)) to calculate how different probability distributions
were for different verbs, they operationalized two types of learning
onset. Between-class onset is declared when a verb's probability
distribution over next-token predictions is more similar to same-class
verbs than to verbs from the opposite class, assessed via a one-tailed
Mann-Whitney U-test. Within-class onset is declared when verbs of the
same class begin to diverge from one another as item-level
idiosyncrasies are learned, measured as the mean within-class
\(D_\text{JS}\). Their key finding was that for to-Dative and Motion
verbs, between-class differentiation emerged approximately 50 training
steps before within-class divergence, consistent with the
abstraction-first account.

To rule out the possibility that the abstraction-first pattern was an
artifact of the training corpus rather than the model,
\citet{jian2026humanstransformerlms} implemented a count-based
distributional baseline using add-\(k\) smoothing applied to
co-occurrence vectors from the same corpus. This baseline showed the
opposite pattern: item-specific divergence emerged before class
distinctions became apparent, matching the exemplar-first prediction.
\citet{jian2026humanstransformerlms} interpret this contrast as evidence
that GPT-2 has an abstraction-first learning bias that the count-based
baseline lacks. This interpretation, however, rests on the assumption
that the onset ordering reliably reflects learning strategy rather than
incidental properties of the learner, such as how sensitive the learner
is to individual observations. Whether the ordering is in fact
diagnostic of an abstraction-first bias, or whether it is a consequence
of learner sensitivity, remains an open question.

\subsection{Present Study}\label{present-study}

We investigate this question through controlled simulations using the
same verb classes, distributional metrics, and onset criteria as
\citet{jian2026humanstransformerlms}, making our results directly
comparable to theirs. We use two idealized learner models that share an
identical data-generation process but differ in how they estimate verb
distributions from observations. Both are \emph{pure memorizers}: each
verb's distribution is estimated entirely from its own observations,
with no information flow between verbs, no class-level representations,
and no capacity to generalize.

The first model, the \textbf{Zero-Sensitivity Learner} (ZSL), provides a
noiseless analytical baseline in which each new observation shifts the
learner's estimate smoothly toward the true distribution, unperturbed by
the idiosyncratic identity of individual tokens. The second model, the
\textbf{Variable-Sensitivity Learner} (VSL), is a Dirichlet-Multinomial
learner that formalizes the add-\(k\) smoothing of
\citet{jian2026humanstransformerlms}'s count-based baseline within a
Bayesian framework, with sensitivity to individual observations
controlled by a single parameter \(k\). At low \(k\), early observations
dominate and idiosyncratic tokens drive the trajectory; at high \(k\),
the estimate is heavily smoothed toward uniformity, suppressing
item-level noise. We use these exemplar models to explore whether
observation-level sensitivity can lead to the appearance of
abstraction-first learning.

Our key contributions are as follows:

\begin{itemize}
\tightlist
\item
  We demonstrate that the ordering of between-class onset (\(o_b\)) and
  within-class onset (\(o_w\)) is not a fixed property of abstract-first
  or exemplar-first learning, but depends on the interaction between
  learner sensitivity (\(k\)) and the structure of the input
  distribution.
\item
  We introduce the Zero-Sensitivity Learner (an exemplar model) as a
  noiseless baseline, showing that \(o_b < o_w\) holds universally when
  observation-level noise is removed.
\item
  We show that the Variable-Sensitivity Learner (also a pure exemplar
  model) replicates the exemplar-first ordering (\(o_w < o_b\)) of the
  count-based baseline at low \(k\) and transitions to the
  abstraction-first ordering (\(o_b < o_w\)) at high \(k\),
  demonstrating that the onset ordering tracks learner sensitivity to
  individual observations rather than abstraction-first or
  exemplar-first learning.
\end{itemize}

\section{Simulations}\label{simulations}

We formalize our central claim through two simulations that share an
identical data-generation process but differ in how the learner
estimates verb distributions from observations. Both models are
\textbf{pure memorizers} (i.e., exemplar models): each verb's
distribution is estimated solely from its own observations, completely
independently of other verbs, with no information flow between verbs and
no class-level representations. Both models assume that the learner
begins from a near-uniform state: before sufficient evidence has
accumulated, all verbs are nearly indistinguishable. We confirm this
assumption empirically for a transformer language model in
Appendix~\ref{sec-app-a}.\footnote{All code and analyses in this paper
  can be found here:
  \url{https://github.com/znhoughton/exemplar-abstraction-sims}.}

\subsection{Data Generation}\label{data-generation}

The following data-generation procedure is identical across both models;
what differs between them is exclusively the learner model: how each
learner constructs its estimate \(\hat{P}_v\) from the available
evidence, described separately for each model below.

\subsubsection{Verb Classes and
Vocabulary}\label{verb-classes-and-vocabulary}

Following \citet{jian2026humanstransformerlms}, we use two verb classes
of 35 and 36 items respectively. The specific verb types are irrelevant
to the simulations; only the class sizes matter. The vocabulary is
simplified to \(V = 1000\) tokens for computational tractability; the
qualitative results are robust to vocabulary size.

\subsubsection{Ground Truth
Distribution}\label{ground-truth-distribution}

Each verb \(v\) has \(n_\text{pref} = 50\) preferred tokens drawn from
the vocabulary which represent the tokens that receive a non-zero
probability after encountering verb \(v\). Two proportion parameters,
class\_overlap \(\in [0, 1]\) and item\_overlap \(\in [0, 1]\), control
how much token sharing occurs across and within classes. Preferred
tokens are partitioned into three types:

\begin{itemize}
\tightlist
\item
  \textbf{Cross tokens}
  (\(n_\text{cross} = \lfloor\text{class\_overlap} \times n_\text{pref}\rfloor\)):
  shared by all verbs in both classes, representing distributional
  properties shared across verb classes.
\item
  \textbf{Within-class tokens}
  (\(n_\text{within} = \lfloor(\text{item\_overlap} - \text{class\_overlap}) \times n_\text{pref}\rfloor\)):
  shared among verbs of the same class only, encoding class-level
  argument preferences.
\item
  \textbf{Idiosyncratic tokens}
  (\(n_\text{idio} = n_\text{pref} - n_\text{cross} - n_\text{within}\)):
  unique to each verb,\footnote{Given the vocabulary size, idiosyncratic
    tokens cannot always be fully unique across all 71 verbs: when
    \((N_A + N_B) \times n_\text{idio}\) exceeds the size of the
    available token pool, some reuse is unavoidable by construction.
    Token assignment is designed to spread this reuse as evenly as
    possible, so that no idiosyncratic token is ever shared by more than
    two verbs.} representing item-specific collocational preferences.
\end{itemize}

Preferred token weights are drawn from a log-normal distribution
\(\text{LogNormal}(\log\mu - \frac{\sigma^2}{2},\, \sigma)\),
parameterized so that \(\mathbb{E}[\text{weight}] = \mu\) regardless of
\(\sigma\).\footnote{For a standard \(\text{LogNormal}(\mu, \sigma)\)
  distribution, the expectation is \(e^{\mu + \frac{\sigma^2}{2}}\)
  rather than \(e^\mu\) due to Jensen's Inequality. By shifting the
  first parameter to \(\log\mu - \frac{\sigma^2}{2}\), the
  \(+\frac{\sigma^2}{2}\) term cancels out, forcing the final
  expectation to be exactly \(\mu\).} Non-preferred tokens receive a
uniform background weight of 1.0. Each verb's true distribution
\(P_v(t)\) (the probability of vocabulary item
\(t \in \{1, \ldots, V\}\)) is obtained by normalizing these raw
weights:

\begin{equation}\phantomsection\label{eq-true-dist}{P_v(t) = \frac{w_t}{\sum_{i=1}^{V} w_i}}\end{equation}

\subsubsection{Onset Criteria}\label{onset-criteria}

Both criteria match \citet{jian2026humanstransformerlms}'s Experiments 1
and 2. Pairwise distances between verb distributions are measured using
the Jensen-Shannon divergence:

\begin{equation}\phantomsection\label{eq-djs}{D_\text{JS}(P, Q) = H\!\left(\tfrac{P+Q}{2}\right) - \tfrac{1}{2}\bigl[H(P) + H(Q)\bigr]}\end{equation}

where \(H(\cdot)\) denotes Shannon entropy. \(D_\text{JS}\) ranges from
0 (identical distributions) to \(\log 2 \approx 0.693\) (maximally
dissimilar distributions).

\textbf{Between-class onset (}\(o_b\)) is declared at the first step at
which \(\geq 10\%\) of verbs individually pass a one-tailed Mann-Whitney
U-test (\(p < 0.001\)) comparing that verb's \(D_\text{JS}\) distances
to out-of-class verbs against its distances to same-class verbs.

\textbf{Within-class onset (}\(o_w\)) is declared at the first step at
which the mean within-class \(D_\text{JS}\), \(\bar{D}_\text{within}\),
exceeds 0.01 for three consecutive steps:

\begin{equation}\phantomsection\label{eq-d-within}{
\begin{split}
\bar{D}_\text{within} &= \frac{1}{2}\left[\frac{1}{\binom{N_A}{2}}\sum_{i<j\,\in\,A} D_\text{JS}(i,j)\right.\\
&\left.\qquad + \frac{1}{\binom{N_B}{2}}\sum_{i<j\,\in\,B} D_\text{JS}(i,j)\right]
\end{split}
}\end{equation}

where \(A\) and \(B\) denote the separate verb classes (\(N_A = 35\) and
\(N_B = 36\) verbs respectively), and the sums are over all unique
within-class verb pairs.

The two criteria are fundamentally asymmetric: \(o_b\) is a
\emph{relative} rank test that fires whenever between-class distances
consistently exceed within-class distances, regardless of absolute
magnitude. \(o_w\) is an \emph{absolute} threshold test that requires
\(D_\text{JS}\) to reach a specific magnitude. As we show below, this
asymmetry is the key mechanism underlying the \(o_b < o_w\) ordering.

All models share the same structural parameters
(\(\mu \in \{10, 30, 60, 100\}\); \(\sigma \in \{0.5, 1.0, 1.5\}\);
item\_overlap \(\in \{0.5, 0.6, 0.7\}\); class\_overlap
\(\in \{0.2, 0.3, 0.4\}\)), subject to the constraint class\_overlap
\(<\) item\_overlap, yielding 108 valid structural combinations (full
details in Appendix~\ref{sec-app-b}). Each model additionally varies its
own sensitivity parameters, which are described in the respective
sections below. All conditions are replicated 50 times per combination.
The models share identical data-generation procedures and onset
criteria; they differ only in how the learner estimates \(\hat{P}_v\)
from available evidence.

\subsection{Zero-Sensitivity Learner}\label{zero-sensitivity-learner}

\subsubsection{Learner Model}\label{learner-model}

The Zero-Sensitivity Learner represents the theoretical limit of a
maximally conservative learner: one whose estimates are entirely
unaffected by any individual token observation. Whereas a realistic
distributional learner is shaped by whichever specific tokens happen to
appear earliest in a verb's contexts (which may be idiosyncratic and
unrepresentative), the Zero-Sensitivity Learner's trajectory is
noiseless and monotone, moving smoothly from complete ignorance toward
full knowledge of each verb's distribution. It is an analytical baseline
that isolates the effect of learning progress from observation-level
noise.

Formally, the learner's estimate of verb \(v\)'s distribution at
learning stage \(\alpha \in [0, 1]\) is a linear interpolation between
the uniform distribution and \(P_v\):

\begin{equation}\phantomsection\label{eq-zsl}{\hat{P}_v(\alpha) = (1 - \alpha) \cdot \text{Uniform} + \alpha \cdot P_v}\end{equation}

Each component has a precise interpretation. \(P_v\) is verb \(v\)'s
\textbf{true distribution}: the stable distribution that verb would
converge to given unlimited data. Critically, each verb has its own
\(P_v\): dative verbs like \emph{give} and \emph{sell} have
systematically different true distributions from motion verbs like
\emph{go} and \emph{run} (reflecting different argument preferences),
and individual verbs within the same class have modestly distinct
distributions from one another (reflecting idiosyncratic collocational
preferences). The \textbf{Uniform} term (\(1/V\) for each of the \(V\)
vocabulary items) represents the learner's starting state: a learner
with no distributional knowledge assigns equal probability to every
token regardless of which verb appears in context.

The parameter \(\alpha\) is the central quantity of this model.
Mathematically, it is the weight placed on the true distribution
relative to the uniform baseline, linearly interpolating between the
two. Conceptually, \(\alpha\) represents the stage of learning: at
\(\alpha = 0\), the learner's estimated distributions are entirely
uniform, reflecting no learned knowledge of verb-specific structure; at
\(\alpha = 1\), the learner has fully converged to each verb's true
distribution \(P_v\). Intermediate values of \(\alpha\) correspond to
intermediate stages of convergence.

Because the trajectory is deterministic, the only signals driving
divergence between verb distributions are the structural properties
encoded in the \(P_v\) themselves: the class-level structure
(within-class shared tokens) creates systematic between-class divergence
detectable by \(o_b\), while idiosyncratic verb-specific tokens create
within-class divergence that drives \(o_w\).

We evaluate the \(71 \times 71\) pairwise \(D_\text{JS}\) matrix and
check both onset criteria at 25 log-spaced values of \(\alpha\) from
0.001 to 1.0, tracing the full learning trajectory. Log spacing
concentrates evaluation points near \(\alpha = 0\), where onset behavior
first emerges. Each parameter combination is run for 50 random seeds,
yielding \(108 \times 50 = 5{,}400\) runs total.

\subsubsection{ZSL Results}\label{zsl-results}

\begin{figure}

\centering{

\pandocbounded{\includegraphics[keepaspectratio]{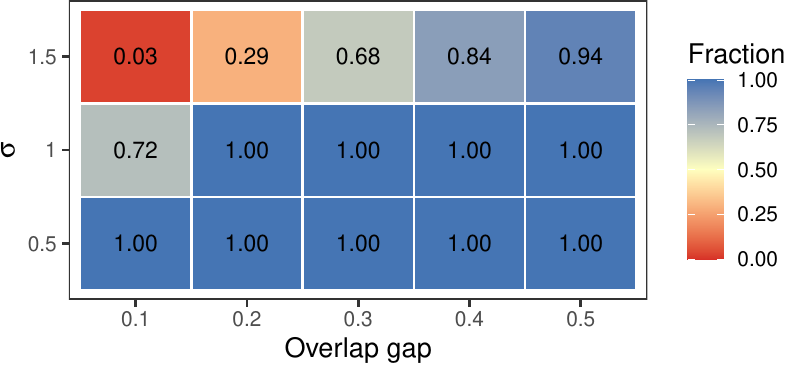}}

}

\caption{\label{fig-m1-heatmap}Zero-Sensitivity Learner: fraction of
seeds with \(o_b < o_w\) across the parameter space, averaged over
\(\mu \in \{10, 30, 60, 100\}\). Blue = \(o_b < o_w\) dominates; red =
\(o_w < o_b\) dominates; yellow \(\approx 0.5\) marks the transition.}

\end{figure}

The Zero-Sensitivity Learner produces \(o_b < o_w\) in 85\% of seeds
overall, \(o_w < o_b\) in 12\%, and neither onset detected in 3\%.
Figure~\ref{fig-m1-heatmap} shows the fraction of seeds yielding
\(o_b < o_w\) across the \((\sigma, \text{gap})\) parameter space,
averaged over \(\mu \in \{10, 30, 60, 100\}\).

The \(o_b < o_w\) result is universal for \(\sigma = 0.5\) across all
parameter combinations. For \(\sigma = 1.0\), the single exception is
the weakest possible class signal (gap \(= 0.1\)), where \(o_w < o_b\)
occurs in 28\% of seeds on average; all other combinations produce
\(o_b < o_w\). For \(\sigma = 1.5\), \(o_b < o_w\) breaks down at small
overlap gaps, where high variance between verbs creates sufficient
idiosyncratic noise to drive \(o_w\) before any class structure
accumulates. As the between-class gap increases, \(o_b < o_w\) recovers
to 95\% at gap \(= 0.5\).

The exceptions are concentrated in the parameter region with the weakest
class signal: small gap, high \(\sigma\), and low \(\mu\) (see
Appendix~\ref{sec-app-c} for the full per-\(\mu\) breakdown). This is
the expected behavior: the \(o_b\) criterion requires a detectable class
signal, and neither ordering can be reliably observed when the class
structure is too weak to distinguish from noise. In the next section, we
explore what might be driving this effect.

\subsection{Variable-Sensitivity
Learner}\label{variable-sensitivity-learner}

\subsubsection{Learner Model}\label{learner-model-1}

The Variable-Sensitivity Learner accumulates distributional evidence
incrementally, constructing its estimate of each verb's distribution
from the tokens observed in that verb's contexts. Unlike the
Zero-Sensitivity Learner, whose trajectory is noiseless and
deterministic, the Variable-Sensitivity Learner's estimates are shaped
by whichever specific tokens appear earliest, including idiosyncratic
tokens that may not reflect the verb's stable class-level distribution.
How strongly those early, potentially unrepresentative observations
shape the learner's estimate is governed entirely by the smoothing
parameter \(k\).

Formally, the model is a \textbf{Dirichlet-Multinomial} conjugate
Bayesian learner. Before any observations, the learner holds a
\textbf{prior} over verb \(v\)'s token distribution: a symmetric
Dirichlet, \(\text{Dir}(k, k, \ldots, k)\), encoding equal prior belief
across all \(V\) token types. This is the multinomial analogue of a
Beta\((\alpha_1, \alpha_2)\) prior: just as Beta\((k, k)\) encodes \(k\)
pseudo-successes and \(k\) pseudo-failures before any data,
\(\text{Dir}(k, \ldots, k)\) encodes \(k\) pseudo-observations of each
of the \(V\) token types before any verb contexts are seen. Larger \(k\)
means a stronger prior: more pseudo-evidence the real data must overcome
before the estimated distribution departs from uniform.

Rather than sweeping \(\alpha\) directly, the Variable-Sensitivity
Learner tracks \(n_\text{obs}\): the number of tokens the learner has
encountered for each verb. The \textbf{likelihood} is Multinomial: each
token is drawn independently from \(P_v\). Because the Dirichlet is the
conjugate prior to the Multinomial, the \textbf{posterior} after
\(n_\text{obs}\) observations, where \(c_t\) denotes the count of
vocabulary item \(t\), is simply:

\begin{equation}\phantomsection\label{eq-dirichlet}{
\begin{split}
&\text{Dir}\!\bigl(k + c_1,\;\ldots,\\
&\qquad k + c_V\bigr)
\end{split}
}\end{equation}

Because the mean of a Dirichlet\((\alpha_1, \ldots, \alpha_V)\)
distribution for token \(t\) is \(\alpha_t / \sum_j \alpha_j\), the
\textbf{posterior mean} for token \(t\) follows directly: substituting
\(\alpha_t = k + \text{count}(t \mid v, n_\text{obs})\) and
\(\sum_j \alpha_j = kV + n_\text{obs}\) gives:

\begin{equation}\phantomsection\label{eq-posterior-mean}{\hat{P}_v(t) = \frac{\text{count}(t \mid v,\, n_\text{obs}) + k}{n_\text{obs} + k \cdot V}}\end{equation}

This is the posterior mean of the Dirichlet-Multinomial, derived as a
direct consequence of Bayesian updating. The numerator adds \(k\)
pseudo-observations to the raw count; the denominator incorporates both
the \(n_\text{obs}\) real observations and the \(k \cdot V\) total
pseudo-observations from the prior.

The parameter \(k\) is therefore the \textbf{concentration} of the
Dirichlet prior, measuring how strongly the learner holds on to its
initial uniform belief. Conceptually, \(k\) controls how
\textbf{conservative} (or equivalently, how \textbf{sensitive}) the
learner is to individual observations. A \textbf{high-}\(k\) learner is
conservative: the large pseudo-count mass \(k \cdot V\) in the
denominator means each new observation shifts the estimate by only a
small amount, requiring many consistent observations before the
estimated distribution departs meaningfully from uniform. This
suppresses idiosyncratic early-observation noise and allows stable,
class-consistent patterns to accumulate gradually. A \textbf{low-}\(k\)
learner is highly sensitive: with few pseudo-counts, each observation
has a large effect on the estimate, and the learner's distribution for a
verb quickly reflects whichever tokens happened to appear first,
regardless of whether those tokens are class-typical or idiosyncratic to
that verb.

As \(k \to \infty\), stochastic deviations around the posterior mean
vanish and the Variable-Sensitivity Learner converges to the
deterministic trajectory of the Zero-Sensitivity Learner (see
Appendix~\ref{sec-app-d} for the derivation): the expected value of
\(\hat{P}_v\) at any \(n_\text{obs}\) is exactly the Zero-Sensitivity
Learner's interpolation with
\(\alpha = n_\text{obs} / (n_\text{obs} + kV)\), and stochastic
deviations around that expectation vanish as \(k\) grows.

In concrete terms, the learner accumulates tokens for all verbs in
parallel. At each observation, every verb \(v\) receives one token: a
vocabulary item drawn from \(P_v\), which increments that item's running
count for \(v\) by 1. Equation~\ref{eq-posterior-mean} can be evaluated
at any point to read off the current estimate \(\hat{P}_v\): each token
type's count inflated by \(k\) pseudo-observations, divided by
\(n_\text{obs} + kV\), where \(n_\text{obs}\) is the number of tokens
accumulated for that verb. We simulate the learner across 5,000 tokens
per verb and examine the learner's state at 20 log-spaced values of
\(n_\text{obs}\) from 1 to 5,000; because the same sequence underlies
all snapshots for a given seed, the tokens at \(n_\text{obs} = 100\) are
a strict superset of those at \(n_\text{obs} = 10\). Every verb
encounters the same number of tokens at every checkpoint, so verb
frequency plays no role; the only source of variability across verbs and
seeds is which specific tokens happen to appear.\footnote{The assumption
  of equal observations per verb is an idealization. In
  \citet{jian2026humanstransformerlms}'s corpus baseline, verbs differ
  substantially in frequency, so more frequent verbs accumulate
  observations faster. We tested a Zipfian extension (Model 3) in which
  the individual observations were sampled from a Zipfian distribution
  whereby each verb's observation count is proportional to its frequency
  rank. The results replicate the qualitative pattern from the
  equal-frequency VSL model; this analysis is presented in
  Appendix~\ref{sec-app-e}.}

\subsubsection{VSL Results}\label{vsl-results}

Overall results by \(k\) are shown in Table~\ref{tbl-m2-summary-table}.
For \(k \leq 0.1\) (effective pseudo-count total
\(k \cdot V \leq 100\)), \(o_w < o_b\) holds in approximately 99\% of
seeds regardless of structural parameters. The transition from
\(o_w < o_b\) to \(o_b < o_w\) begins at \(k = 0.5\)
(\(k \cdot V = 500\) pseudo-counts), where both orderings occur with
roughly equal frequency and many seeds produce a tie (\(o_b = o_w\) at
the same \(n_\text{obs}\) grid point). By \(k = 1.0\)
(\(k \cdot V = 1{,}000\) pseudo-counts), the prior is strong enough to
fully suppress idiosyncratic early-observation noise: \(o_b < o_w\)
predominates across the majority of structural parameter combinations.

\begin{table}[H]

\centering{

\begin{tabular}{rrrrr}
\toprule
$k$ & $o_b < o_w$ & $o_w < o_b$ & Tie & Neither\\
\midrule
0.001 & 0.000 & 0.990 & 0.000 & 0.010\\
0.010 & 0.000 & 0.989 & 0.000 & 0.011\\
0.100 & 0.000 & 0.983 & 0.003 & 0.015\\
0.500 & 0.403 & 0.406 & 0.167 & 0.024\\
1.000 & 0.703 & 0.170 & 0.099 & 0.028\\
\bottomrule
\end{tabular}

}

\caption{\label{tbl-m2-summary-table}Variable-Sensitivity Learner
overall means by \(k\). `Tie' indicates both onsets detected at the same
\(n_{\text{obs}}\) grid point.}

\end{table}%

\begin{figure}

\centering{

\includegraphics[width=1\linewidth,height=\textheight,keepaspectratio]{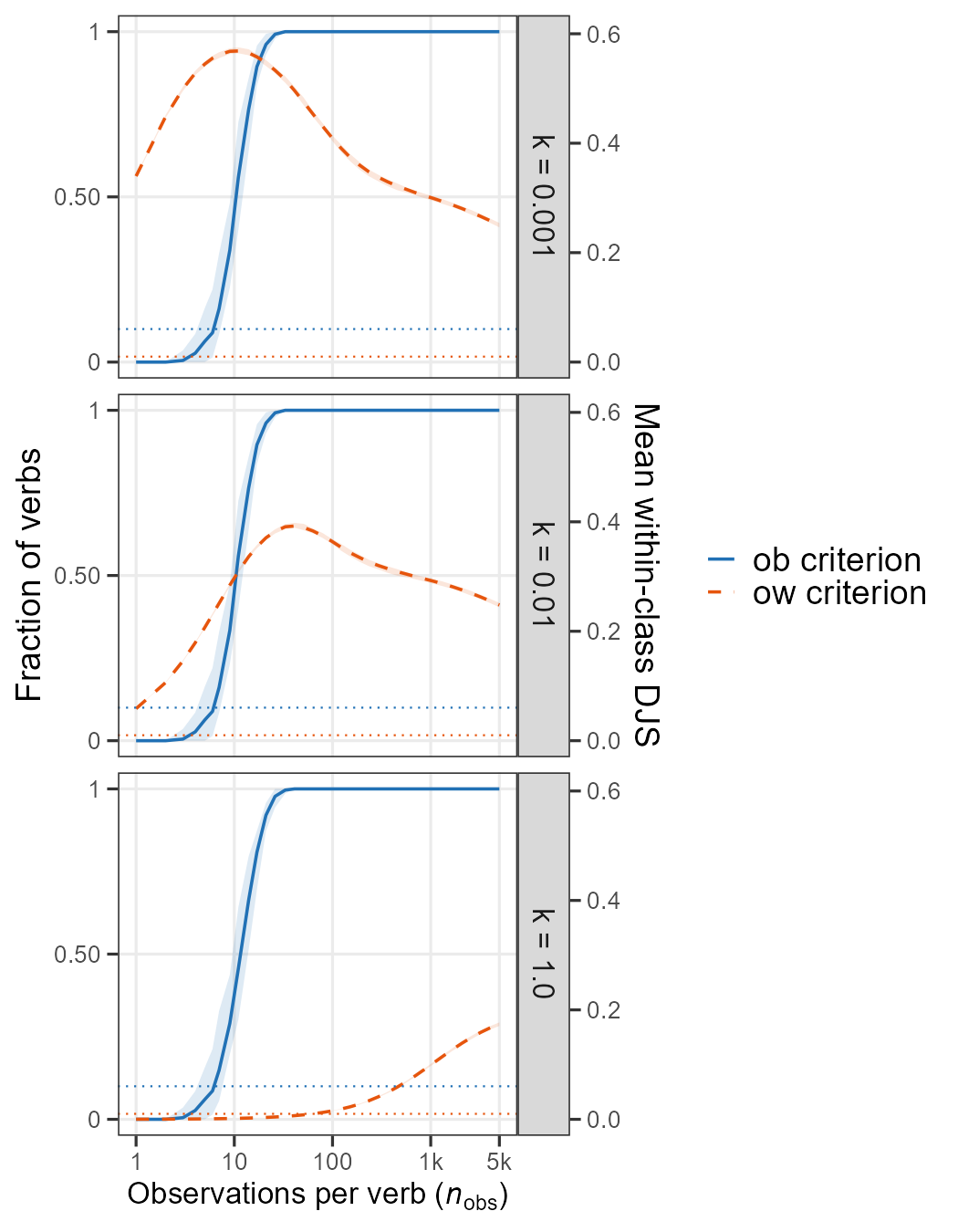}

}

\caption{\label{fig-k-trajectories}Variable-Sensitivity Learner: \(o_b\)
and \(o_w\) criteria over observations per verb (\(n_\text{obs}\)) for
three values of \(k\) straddling the transition. Blue solid: \(o_b\)
criterion (fraction of verbs passing the Mann-Whitney test, left axis).
Red dashed: \(o_w\) criterion (mean within-class \(D_\text{JS}\), right
axis). Dotted lines mark the onset thresholds (CLASS\_FRAC \(= 0.10\);
DJS\_THRESH \(= 0.01\)). Parameters: \(\mu = 100\), \(\sigma = 0.5\),
item\_overlap \(= 0.7\), class\_overlap \(= 0.2\); 20 seeds; 95\% CI
ribbons.}

\end{figure}

Figure~\ref{fig-k-trajectories} illustrates why: it shows the \(o_b\)
and \(o_w\) criterion trajectories over \(n_\text{obs}\) for three
values of \(k\) straddling the transition, making the mechanism directly
visible. At low \(k\), \(o_w\) spikes immediately at the first
observations while \(o_b\) accumulates slowly; at high \(k\), \(o_w\) is
suppressed and \(o_b\) detects the class-level asymmetry first.

\begin{figure}

\centering{

\pandocbounded{\includegraphics[keepaspectratio]{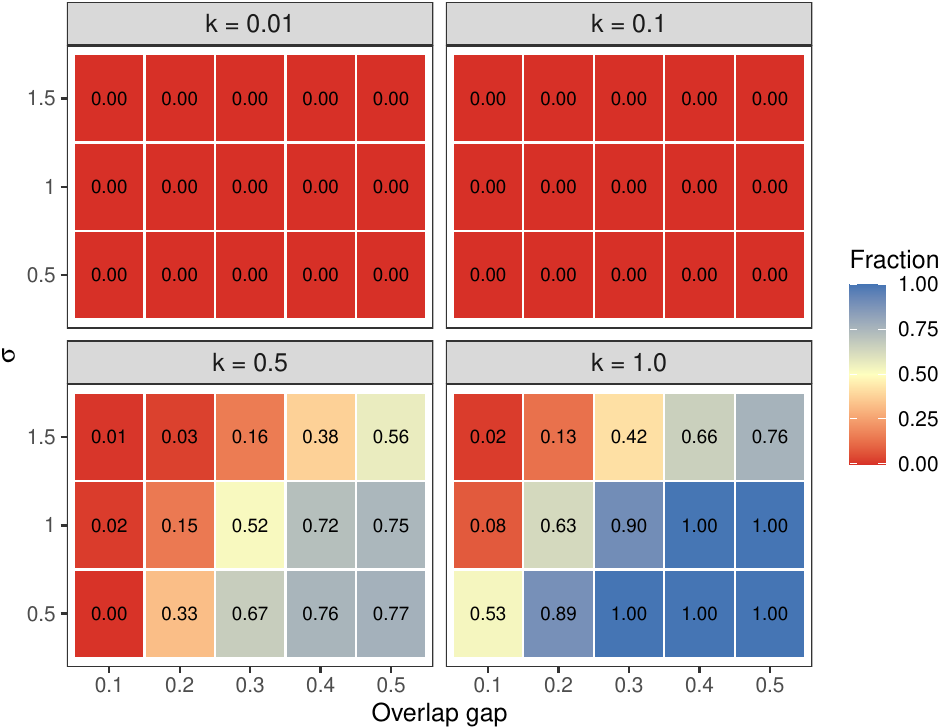}}

}

\caption{\label{fig-m2-heatmap}Variable-Sensitivity Learner: fraction of
seeds with \(o_b < o_w\) by \(k\), \(\sigma\), and overlap gap, averaged
over \(\mu \in \{10, 30, 60, 100\}\). Blue = \(o_b < o_w\) dominates;
red = \(o_w < o_b\) dominates. At \(k \leq 0.1\) virtually all seeds
produce \(o_w < o_b\); the transition emerges at \(k = 0.5\) and
\(o_b < o_w\) dominates by \(k = 1.0\).}

\end{figure}

The breakdown across the full \((\sigma, \text{gap})\) parameter space
is shown in Figure~\ref{fig-m2-heatmap} for
\(k \in \{0.01, 0.1, 0.5, 1.0\}\) (results at \(k = 0.001\) are
virtually identical to \(k = 0.01\) and are omitted from the figure;
full numerical results for all \(k\) values in
Appendix~\ref{sec-app-f}). At \(k = 0.01\), \(o_w < o_b\) is universal
across nearly all parameter combinations, with a small number of seeds
failing to detect either onset only in the weak-signal corner (small
gap, high \(\sigma\), low \(\mu\)). At \(k = 0.1\), the pattern is
qualitatively unchanged: \(o_w < o_b\) remains near-universal,
indicating that \(k \cdot V = 100\) pseudo-counts are still insufficient
to suppress idiosyncratic noise. The transition emerges at \(k = 0.5\)
and \(o_b < o_w\) dominates by \(k = 1.0\).

The \(o_w < o_b\) result is completely invariant to the \(D_\text{JS}\)
threshold at all \(k\) values (see Appendix~\ref{sec-app-g}). At low
\(k\), within-class \(D_\text{JS}\) spikes immediately to values
exceeding any reasonable threshold in the tested range, so \(o_w\) fires
at the first observation regardless of threshold choice. The
\(o_b < o_w\) result in the high-\(k\) regime is similarly robust across
all tested values of CLASS\_FRAC and \(D_\text{JS}\) threshold. The
qualitative pattern also holds when verbs accumulate counts at rates
proportional to their frequency rather than equally, relaxing what may
be an unrealistic assumption (see Appendix~\ref{sec-app-e}).

Taken together, both learners demonstrate that the \(o_b < o_w\)
ordering is a predictable consequence of low sensitivity to individual
observations, not of an abstraction-first inductive bias. Both are pure
memorizers: each verb's distribution is estimated entirely from its own
smoothed token counts, with no information flow between verbs and no
class-level representations. Despite this, both produce \(o_b < o_w\)
whenever sensitivity is sufficiently low: at the deterministic limit
(Zero-Sensitivity Learner) or with sufficient add-\(k\) smoothing
(Variable-Sensitivity Learner, high \(k\)). The ordering of onsets is
therefore diagnostic of observation sensitivity, not of abstraction.
Conversely, high sensitivity (low \(k\)) produces \(o_w < o_b\), which
is the signature of a learner shaped by idiosyncratic early observations
before consistent class-level patterns accumulate.

\section{Discussion and Conclusion}\label{discussion-and-conclusion}

The present study demonstrates that the abstraction-first ordering
arises in pure memorizers with no class-level representations, no
information flow between verbs, and no inductive bias toward
generalization, provided that sensitivity to individual observations is
sufficiently low. The onset ordering is therefore not diagnostic of
abstraction-first learning; it reflects how sensitive the learner is to
individual observations. These results do not preclude the possibility
that transformer models are abstraction-first learners, but they show
that the evidence \citet{jian2026humanstransformerlms} provide is
insufficient to establish this: a high-\(k\) exemplar memorizer produces
the same apparent onset pattern. There is independent reason to think
transformers may behave like high-\(k\) learners:
\citet{kapatsinski2026transformersperformadaptive} shows that
transformers perform adaptive partial pooling, weighting individual
observations against class-level averages in a frequency-sensitive way.
This suppresses idiosyncratic noise in the same way high \(k\) does,
providing a mechanistic basis for why GPT-2 might show apparent
abstraction-first ordering. To be clear, this pooling should not itself
be taken as evidence of abstraction: pooling over instances can be a
natural consequence of how distributed representations are structured,
reflecting the geometry of the representational space rather than the
presence of discrete class-level structures separable from item-specific
knowledge. The \(o_b < o_w\) pattern could arise from this distributed
geometry without any explicit encoding of class membership.

A deeper challenge concerns whether the exemplar-versus-abstraction
dichotomy is well-defined for distributed representations in the first
place. The \(o_b\) and \(o_w\) onset criteria presuppose that
between-class differentiation and within-class divergence are separable
learning events, with one marking the emergence of abstract class
knowledge and the other the accumulation of item-specific exemplars. But
for a learner whose representations are purely distributional, these may
not be separable. Learning the distributional properties of \emph{give},
which argument types and constructions it favors, necessarily encodes
its class membership, because what makes \emph{give}'s distribution
distinctive partly consists of the features it shares with \emph{sell}
and that set it apart from \emph{go}. The class-level signal is not a
separate layer of knowledge sitting above the item-specific knowledge;
it is embedded within it. The onset criteria treat between-class and
within-class divergence as tracking two separable learning events, but
for distributed representations there may be no clean seam between the
two, and interpreting them as indexing qualitatively distinct kinds of
knowledge assumes a conceptual distinction that the representations
themselves may not support.

In summary, the ordering of between-class and within-class onset is a
function of observation sensitivity, not of learning strategy. A pure
memorizer produces the abstraction-first pattern whenever sensitivity is
sufficiently low, which means that
\citet{jian2026humanstransformerlms}'s methodology cannot distinguish
abstraction-first learning from high-\(k\) exemplar learning. More
broadly, the exemplar-versus-abstraction distinction may itself be
ill-defined for distributed representations: a word's class-level
properties are embedded within its item-specific distribution, making
the two difficult to dissociate from distributional evidence alone.
Whether transformer models are abstraction-first learners in any
theoretically meaningful sense remains an open question, but answering
it will require methodology that goes beyond the ordering of onset
criteria.

\section{Limitations}\label{limitations}

The present simulations use synthetic verb distributions rather than
real corpus frequencies. This is a deliberate design choice that allows
precise control over the overlap structure, but it means the specific
parameter values that produce each ordering do not map directly to
naturalistic language. This concern is mitigated by the breadth of the
parameter sweep: the core results hold across wide ranges of \(\mu\),
\(\sigma\), item overlap, and class overlap, suggesting they are not
artifacts of any particular distributional assumption. Similarly, we use
two verb classes rather than the four analyzed by
\citet{jian2026humanstransformerlms}; a multi-class scenario introduces
additional between-class pairings that our model does not capture. Given
that the onset ordering is driven by the asymmetry between relative and
absolute criteria rather than by the number of classes, we do not expect
this to change the qualitative conclusions, but it remains to be
verified.

\bibliography{references}
\renewcommand{\bibliography}[1]{}
\clearpage
\onecolumn
\appendix

\section{Empirical Confirmation of Near-Uniform Starting
State}\label{sec-app-a}

Both simulations assume that the learner begins from a state where all
verb distributions are near-uniform: the model initially assigns
essentially equal probability to all tokens regardless of which verb
appears in the context. This is the \(\alpha = 0\) state in the
Zero-Sensitivity Learner and the \(n_\text{obs} \approx 0\) state in the
Variable-Sensitivity Learner.

To confirm this assumption for an actual transformer language model, we
computed next-token distributions for all 71 J\&M verbs at training
checkpoint 1 of \texttt{znhoughton/opt-babylm-125m-1ep-dense-seed964}
(OPT-125M trained on 100M words of BabyLM data). We used the unified
template \texttt{"They\ \{verb\}\ to\ the"} for both verb classes. Using
a single template is essential: class-specific templates would introduce
between-class \(D_\text{JS}\) through template differences rather than
verb identity.

\begin{table}[H]

\centering{

\begin{tabular}{ll}
\toprule
Metric & Value\\
\midrule
Template & "They \{verb\} to the"\\
Vocabulary size & 8,192\\
Maximum entropy (uniform) & 9.011 nats\\
Mean verb entropy at checkpoint 1 & 8.857 nats\\
Fraction of maximum entropy & 98.3\%\\
Mean within-class $D_\text{JS}$ & 0.0074\\
Mean between-class $D_\text{JS}$ & 0.0074\\
Between / within ratio & 1.002\\
\bottomrule
\end{tabular}

}

\caption{\label{tbl-babylm}Next-token distribution statistics at
checkpoint 1 of OPT-125M (BabyLM). All 71 verbs are at 98.3\% of maximum
entropy at the first checkpoint.}

\end{table}%

At checkpoint 1, all 71 verbs are at 98.3\% of maximum entropy.
Within-class and between-class \(D_\text{JS}\) are virtually identical
(\(\text{ratio} = 1.002\)) and both fall well below the \(o_w\)
threshold of 0.01. There is no class-level signal at step 1; the model
treats all 71 verbs as effectively equivalent. This confirms the
near-uniform starting condition assumed by the simulations.

We note two limitations. First, this confirms the starting condition for
OPT-125M trained on BabyLM data, not for GPT-2 on OpenWebText as in
J\&M. However, the near-uniform starting condition is a general property
of randomly initialized transformers trained with standard
autoregressive objectives: before sufficient gradient updates, all token
predictions default to the marginal distribution of the training data,
independent of verb identity. Second, the OPT-125M vocabulary has 8,192
tokens rather than the GPT-2 vocabulary of 50,257, so the absolute
entropy values are not directly comparable, but the 98.3\% saturation
result is vocabulary-size independent.

\newpage{}

\section{Parameter Grid}\label{sec-app-b}

\begin{table}[H]

\centering{

\begin{tabular}{lll}
\toprule
Parameter & Values tested & Description\\
\midrule
$\mu$ & 10, 30, 60, 100 & Mean lognormal weight on preferred tokens\\
$\sigma$ & 0.5, 1.0, 1.5 & Log-normal spread across preferred tokens\\
item\_overlap & 0.5, 0.6, 0.7 & Within-class token sharing fraction\\
class\_overlap & 0.2, 0.3, 0.4 & Cross-class token sharing fraction\\
Seeds & 50 & Random seeds per parameter combination\\
$k$ (Variable-Sensitivity only) & 0.001, 0.01, 0.1, 0.5, 1.0 & Smoothing strength (controls sensitivity)\\
\bottomrule
\end{tabular}

}

\caption{\label{tbl-params}Parameter grid. All combinations with
class\_overlap \(<\) item\_overlap are valid (108 structural
combinations).}

\end{table}%

Constraint: class\_overlap \(<\) item\_overlap (108 valid structural
combinations). The Zero-Sensitivity Learner is run over the 108
structural combinations only (\(108 \times 50 = 5{,}400\) runs). The
Variable-Sensitivity Learner additionally sweeps \(k\), giving
\(108 \times 5 \times 50 = 27{,}000\) total runs; results at
\(k = 0.001\) are excluded from main-text figures as they are virtually
identical to \(k = 0.01\).

\newpage{}

\section{\texorpdfstring{Detailed Results by
\(\mu\)}{Detailed Results by \textbackslash mu}}\label{sec-app-c}

The following tables show the per-\(\mu\) breakdown for the
Zero-Sensitivity Learner (the results most sensitive to structural
parameters). Each cell is averaged over the (item\_overlap,
class\_overlap) pairs that yield the given overlap gap. \(\sigma = 0.5\)
is omitted: \(o_b < o_w = 1.00\) for all 20 combinations at
\(\sigma = 0.5\) regardless of \(\mu\).

\begin{longtable}{rrlll}

\toprule
gap & $\mu$ & $o_b < o_w$ & $o_w < o_b$ & Neither\\
\midrule
\endfirsthead
\multicolumn{5}{@{}l}{\textit{(continued)}}\\
\toprule
gap & $\mu$ & $o_b < o_w$ & $o_w < o_b$ & Neither\\
\midrule
\endhead

\endfoot
\endlastfoot
0.1 & 10 & 0.24 [0.14, 0.37] & 0.74 [0.56, 0.81] & 0.02\\
0.1 & 30 & 0.72 [0.58, 0.83] & 0.28 [0.04, 0.21] & 0.00\\
0.1 & 60 & 0.92 [0.81, 0.97] & 0.08 [0.00, 0.07] & 0.00\\
0.1 & 100 & 1.00 [0.93, 1.00] & 0.00 [0.00, 0.07] & 0.00\\
0.2 & 10 & 1.00 [0.96, 1.00] & 0.00 [0.00, 0.04] & 0.00\\
0.2 & 30 & 1.00 [0.96, 1.00] & 0.00 [0.00, 0.04] & 0.00\\
0.2 & 60 & 1.00 [0.96, 1.00] & 0.00 [0.00, 0.04] & 0.00\\
0.2 & 100 & 1.00 [0.96, 1.00] & 0.00 [0.00, 0.04] & 0.00\\
0.3 & 10 & 1.00 [0.98, 1.00] & 0.00 [0.00, 0.02] & 0.00\\
0.3 & 30 & 1.00 [0.98, 1.00] & 0.00 [0.00, 0.02] & 0.00\\
0.3 & 60 & 1.00 [0.98, 1.00] & 0.00 [0.00, 0.02] & 0.00\\
0.3 & 100 & 1.00 [0.98, 1.00] & 0.00 [0.00, 0.02] & 0.00\\
0.4 & 10 & 1.00 [0.96, 1.00] & 0.00 [0.00, 0.04] & 0.00\\
0.4 & 30 & 1.00 [0.96, 1.00] & 0.00 [0.00, 0.04] & 0.00\\
0.4 & 60 & 1.00 [0.96, 1.00] & 0.00 [0.00, 0.04] & 0.00\\
0.4 & 100 & 1.00 [0.96, 1.00] & 0.00 [0.00, 0.04] & 0.00\\
0.5 & 10 & 1.00 [0.93, 1.00] & 0.00 [0.00, 0.07] & 0.00\\
0.5 & 30 & 1.00 [0.93, 1.00] & 0.00 [0.00, 0.07] & 0.00\\
0.5 & 60 & 1.00 [0.93, 1.00] & 0.00 [0.00, 0.07] & 0.00\\
0.5 & 100 & 1.00 [0.93, 1.00] & 0.00 [0.00, 0.07] & 0.00\\*
\bottomrule

\caption{\label{tbl-m1-sigma1}Zero-Sensitivity Learner,
\(\sigma = 1.0\): results by overlap gap and \(\mu\). Bracketed values
are 95\% Wilson CIs, pooled over all seeds contributing to each cell.}

\tabularnewline
\end{longtable}

\begin{longtable}{rrlll}

\toprule
gap & $\mu$ & $o_b < o_w$ & $o_w < o_b$ & Neither\\
\midrule
\endfirsthead
\multicolumn{5}{@{}l}{\textit{(continued)}}\\
\toprule
gap & $\mu$ & $o_b < o_w$ & $o_w < o_b$ & Neither\\
\midrule
\endhead

\endfoot
\endlastfoot
0.1 & 10 & 0.00 [0.00, 0.07] & 0.04 [0.01, 0.13] & 0.96\\
0.1 & 30 & 0.00 [0.00, 0.07] & 0.28 [0.17, 0.42] & 0.72\\
0.1 & 60 & 0.04 [0.01, 0.13] & 0.86 [0.71, 0.92] & 0.10\\
0.1 & 100 & 0.08 [0.03, 0.19] & 0.92 [0.79, 0.96] & 0.00\\
0.2 & 10 & 0.03 [0.01, 0.08] & 0.36 [0.25, 0.43] & 0.61\\
0.2 & 30 & 0.18 [0.12, 0.27] & 0.82 [0.66, 0.82] & 0.00\\
0.2 & 60 & 0.37 [0.28, 0.47] & 0.63 [0.32, 0.51] & 0.00\\
0.2 & 100 & 0.60 [0.50, 0.69] & 0.40 [0.13, 0.29] & 0.00\\
0.3 & 10 & 0.11 [0.07, 0.17] & 0.87 [0.74, 0.87] & 0.01\\
0.3 & 30 & 0.65 [0.57, 0.72] & 0.35 [0.07, 0.17] & 0.00\\
0.3 & 60 & 0.96 [0.92, 0.98] & 0.04 [0.00, 0.05] & 0.00\\
0.3 & 100 & 1.00 [0.98, 1.00] & 0.00 [0.00, 0.02] & 0.00\\
0.4 & 10 & 0.37 [0.28, 0.47] & 0.63 [0.39, 0.59] & 0.00\\
0.4 & 30 & 1.00 [0.96, 1.00] & 0.00 [0.00, 0.04] & 0.00\\
0.4 & 60 & 1.00 [0.96, 1.00] & 0.00 [0.00, 0.04] & 0.00\\
0.4 & 100 & 1.00 [0.96, 1.00] & 0.00 [0.00, 0.04] & 0.00\\
0.5 & 10 & 0.78 [0.65, 0.87] & 0.22 [0.01, 0.13] & 0.00\\
0.5 & 30 & 1.00 [0.93, 1.00] & 0.00 [0.00, 0.07] & 0.00\\
0.5 & 60 & 1.00 [0.93, 1.00] & 0.00 [0.00, 0.07] & 0.00\\
0.5 & 100 & 1.00 [0.93, 1.00] & 0.00 [0.00, 0.07] & 0.00\\*
\bottomrule

\caption{\label{tbl-m1-sigma15}Zero-Sensitivity Learner,
\(\sigma = 1.5\): results by overlap gap and \(\mu\). Bracketed values
are 95\% Wilson CIs, pooled over all seeds contributing to each cell.}

\tabularnewline
\end{longtable}

The \(\mu\) effect is most pronounced at \(\sigma = 1.5\) with small
overlap gaps: higher \(\mu\) consistently shifts results toward
\(o_b < o_w\) by making preferred tokens more dominant over background
noise, strengthening the detectable class signal. At \(\sigma = 1.0\),
only the weakest class signal (gap \(= 0.1\), low \(\mu\)) deviates from
universal \(o_b < o_w\).

For the Variable-Sensitivity Learner at \(k = 0.5\) (the transitional
regime where \(\mu\) effects are most relevant), the same \(\mu\)-driven
pattern emerges: in general, higher \(\mu\) and larger gap shift results
toward \(o_b < o_w\), consistent with the interpretation that \(\mu\)
amplifies the class signal relative to idiosyncratic noise (see
Appendix~\ref{sec-app-d} for the formal ZSL-VSL correspondence). Full
per-\(\mu\) results for \(k = 0.5\) are shown below (\(\sigma = 0.5\)
omitted: \(o_b < o_w \approx 1.00\) throughout).

\begin{longtable}{rrrlll}

\toprule
$\sigma$ & gap & $\mu$ & $o_b < o_w$ & $o_w < o_b$ & Neither\\
\midrule
\endfirsthead
\multicolumn{6}{@{}l}{\textit{(continued)}}\\
\toprule
$\sigma$ & gap & $\mu$ & $o_b < o_w$ & $o_w < o_b$ & Neither\\
\midrule
\endhead

\endfoot
\endlastfoot
1.0 & 0.1 & 10 & 0.00 [0.00, 0.07] & 0.98 [0.90, 1.00] & 0.02\\
1.0 & 0.1 & 30 & 0.00 [0.00, 0.07] & 0.98 [0.90, 1.00] & 0.00\\
1.0 & 0.1 & 60 & 0.04 [0.01, 0.13] & 0.86 [0.74, 0.93] & 0.00\\
1.0 & 0.1 & 100 & 0.04 [0.01, 0.13] & 0.84 [0.71, 0.92] & 0.00\\
1.0 & 0.2 & 10 & 0.00 [0.00, 0.04] & 1.00 [0.96, 1.00] & 0.00\\
1.0 & 0.2 & 30 & 0.03 [0.01, 0.08] & 0.71 [0.61, 0.79] & 0.00\\
1.0 & 0.2 & 60 & 0.20 [0.13, 0.29] & 0.31 [0.23, 0.41] & 0.00\\
1.0 & 0.2 & 100 & 0.38 [0.29, 0.48] & 0.18 [0.12, 0.27] & 0.00\\
1.0 & 0.3 & 10 & 0.02 [0.01, 0.06] & 0.95 [0.91, 0.98] & 0.00\\
1.0 & 0.3 & 30 & 0.33 [0.26, 0.41] & 0.09 [0.05, 0.14] & 0.00\\
1.0 & 0.3 & 60 & 0.79 [0.72, 0.85] & 0.00 [0.00, 0.02] & 0.00\\
1.0 & 0.3 & 100 & 0.93 [0.88, 0.96] & 0.00 [0.00, 0.02] & 0.00\\
1.0 & 0.4 & 10 & 0.02 [0.01, 0.07] & 0.83 [0.74, 0.89] & 0.00\\
1.0 & 0.4 & 30 & 0.87 [0.79, 0.92] & 0.00 [0.00, 0.04] & 0.00\\
1.0 & 0.4 & 60 & 0.99 [0.95, 1.00] & 0.00 [0.00, 0.04] & 0.00\\
1.0 & 0.4 & 100 & 1.00 [0.96, 1.00] & 0.00 [0.00, 0.04] & 0.00\\
1.0 & 0.5 & 10 & 0.00 [0.00, 0.07] & 0.60 [0.46, 0.72] & 0.00\\
1.0 & 0.5 & 30 & 1.00 [0.93, 1.00] & 0.00 [0.00, 0.07] & 0.00\\
1.0 & 0.5 & 60 & 1.00 [0.93, 1.00] & 0.00 [0.00, 0.07] & 0.00\\
1.0 & 0.5 & 100 & 1.00 [0.93, 1.00] & 0.00 [0.00, 0.07] & 0.00\\
1.5 & 0.1 & 10 & 0.00 [0.00, 0.07] & 0.08 [0.03, 0.19] & 0.92\\
1.5 & 0.1 & 30 & 0.02 [0.00, 0.10] & 0.40 [0.28, 0.54] & 0.56\\
1.5 & 0.1 & 60 & 0.02 [0.00, 0.10] & 0.94 [0.84, 0.98] & 0.04\\
1.5 & 0.1 & 100 & 0.00 [0.00, 0.07] & 1.00 [0.93, 1.00] & 0.00\\
1.5 & 0.2 & 10 & 0.00 [0.00, 0.04] & 0.51 [0.41, 0.61] & 0.48\\
1.5 & 0.2 & 30 & 0.01 [0.00, 0.05] & 0.95 [0.89, 0.98] & 0.00\\
1.5 & 0.2 & 60 & 0.03 [0.01, 0.08] & 0.89 [0.81, 0.94] & 0.00\\
1.5 & 0.2 & 100 & 0.07 [0.03, 0.14] & 0.78 [0.69, 0.85] & 0.00\\
1.5 & 0.3 & 10 & 0.01 [0.00, 0.04] & 0.96 [0.92, 0.98] & 0.02\\
1.5 & 0.3 & 30 & 0.09 [0.05, 0.14] & 0.81 [0.74, 0.86] & 0.00\\
1.5 & 0.3 & 60 & 0.20 [0.14, 0.27] & 0.47 [0.39, 0.55] & 0.00\\
1.5 & 0.3 & 100 & 0.34 [0.27, 0.42] & 0.24 [0.18, 0.31] & 0.00\\
1.5 & 0.4 & 10 & 0.03 [0.01, 0.08] & 0.94 [0.88, 0.97] & 0.00\\
1.5 & 0.4 & 30 & 0.18 [0.12, 0.27] & 0.40 [0.31, 0.50] & 0.00\\
1.5 & 0.4 & 60 & 0.58 [0.48, 0.67] & 0.04 [0.02, 0.10] & 0.00\\
1.5 & 0.4 & 100 & 0.72 [0.63, 0.80] & 0.02 [0.01, 0.07] & 0.00\\
1.5 & 0.5 & 10 & 0.00 [0.00, 0.07] & 0.96 [0.87, 0.99] & 0.00\\
1.5 & 0.5 & 30 & 0.38 [0.26, 0.52] & 0.06 [0.02, 0.16] & 0.00\\
1.5 & 0.5 & 60 & 0.88 [0.76, 0.94] & 0.00 [0.00, 0.07] & 0.00\\
1.5 & 0.5 & 100 & 1.00 [0.93, 1.00] & 0.00 [0.00, 0.07] & 0.00\\*
\bottomrule

\caption{\label{tbl-m2-k05}Variable-Sensitivity Learner, \(k = 0.5\):
results by \(\sigma\), overlap gap, and \(\mu\). Bracketed values are
95\% Wilson CIs, pooled over all seeds contributing to each cell.}

\tabularnewline
\end{longtable}

\newpage{}

\section{\texorpdfstring{The Zero-Sensitivity Learner as the
\(k \to \infty\) Limit of the Variable-Sensitivity
Learner}{The Zero-Sensitivity Learner as the k \textbackslash to \textbackslash infty Limit of the Variable-Sensitivity Learner}}\label{sec-app-d}

In the Variable-Sensitivity Learner, the smoothed empirical distribution
at corpus size \(n_\text{obs}\) is:

\[\hat{P}_v(t) = \frac{\text{count}(t \mid v,\, n_\text{obs}) + k}{n_\text{obs} + k \cdot V}\]

Taking the expectation over random token draws from \(P_v\), and noting
that
\(\mathbb{E}[\text{count}(t \mid v, n_\text{obs})] = n_\text{obs} \cdot P_v(t)\):

\[\mathbb{E}[\hat{P}_v(t)] = \frac{n_\text{obs} \cdot P_v(t) + k}{n_\text{obs} + k \cdot V} = \frac{n_\text{obs}}{n_\text{obs} + kV} \cdot P_v(t) + \frac{kV}{n_\text{obs} + kV} \cdot \frac{1}{V}\]

Defining \(\alpha(n_\text{obs}) = n_\text{obs} / (n_\text{obs} + kV)\),
this is exactly:

\[\mathbb{E}[\hat{P}_v(n_\text{obs})] = \alpha \cdot P_v + (1 - \alpha) \cdot \text{Uniform}\]

which is Equation~\ref{eq-zsl}, the Zero-Sensitivity Learner's
interpolation formula, with learning progress \(\alpha\).

\newpage{}

\section{Zipfian Verb Frequency Extension (Model 3)}\label{sec-app-e}

The Variable-Sensitivity Learner assumes all verbs are observed equally
often at every checkpoint. Model 3 relaxes this by allocating
observations across verbs according to a Zipfian distribution: a total
budget of \(N_\text{total}\) tokens is drawn, and at each draw a verb is
sampled with probability proportional to \(1/r^s\) (where \(r\) is its
randomly assigned frequency rank and \(s\) is the Zipf exponent),
followed by sampling a single token from that verb's true distribution
\(P_v\). Checkpoints are defined on the \(N_\text{total}\) axis; at each
checkpoint, different verbs have accumulated different numbers of
observations depending on their rank. The mean observations per verb at
the maximum checkpoint equals 5,000, matching Model 2's maximum
\(n_\text{obs}\).

\begin{figure}

\centering{

\pandocbounded{\includegraphics[keepaspectratio]{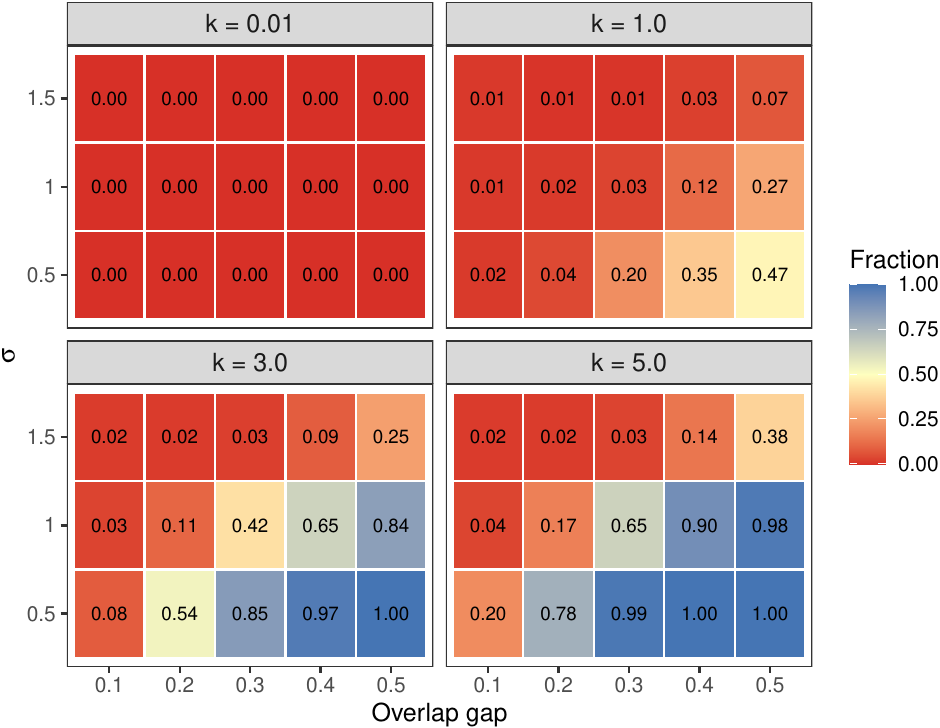}}

}

\caption{\label{fig-m3-heatmap}Zipfian Variable-Sensitivity Learner:
fraction of seeds with \(o_b < o_w\) by \(k\), \(\sigma\), and overlap
gap, averaged over \(\mu \in \{10, 30, 60, 100\}\) and Zipf exponent
\(s \in \{0.5, 1.0, 1.5\}\). Blue = \(o_b < o_w\) dominates; red =
\(o_w < o_b\) dominates. Compare to the equal-frequency
Variable-Sensitivity Learner (Figure~\ref{fig-m2-heatmap}): the
transition shifts to substantially higher \(k\).}

\end{figure}

The qualitative pattern from Model 2 is preserved: as \(k\) increases,
the fraction of seeds showing \(o_b < o_w\) rises and the fraction
showing \(o_w < o_b\) falls. At \(k = 0.01\), virtually all detected
seeds show \(o_w < o_b\) (94\%), while at \(k = 5\), \(o_b < o_w\) is
the dominant outcome (51\% vs.~25\%). The key difference from Model 2 is
that the crossover is shifted to higher \(k\): in the equal-frequency
learner, \(k = 0.5\) already reaches the transition point, whereas in
the Zipfian model it falls around \(k = 3\)--\(5\). This shift arises
because rare verbs accumulate few observations regardless of total input
size, leaving their item-specific distributions heavily smoothed toward
the background until \(k\) is large enough to overcome this. Within the
Zipfian model, steeper frequency skew (\(s = 1.5\)) further delays the
\(o_b\) onset relative to shallower skew (\(s = 0.5\)), because rare
verbs require disproportionately more total tokens to reveal
item-specific structure. Figure~\ref{fig-m3-heatmap} shows this
transition across the full \((\sigma, \text{gap})\) parameter space,
averaged over \(\mu\) and \(s\); the underlying numerical values (50
seeds per combination) are shown in Table~\ref{tbl-model3-full}.

\begin{longtable}{rrrll}

\toprule
$k$ & $\sigma$ & gap & $o_b < o_w$ & $o_w < o_b$\\
\midrule
\endfirsthead
\multicolumn{5}{@{}l}{\textit{(continued)}}\\
\toprule
$k$ & $\sigma$ & gap & $o_b < o_w$ & $o_w < o_b$\\
\midrule
\endhead

\endfoot
\endlastfoot
0.01 & 0.5 & 0.1 & 0.00 [0.00, 0.01] & 0.86 [0.83, 0.89]\\
0.01 & 0.5 & 0.2 & 0.00 [0.00, 0.00] & 0.97 [0.96, 0.98]\\
0.01 & 0.5 & 0.3 & 0.00 [0.00, 0.00] & 1.00 [1.00, 1.00]\\
0.01 & 0.5 & 0.4 & 0.00 [0.00, 0.00] & 1.00 [0.99, 1.00]\\
0.01 & 0.5 & 0.5 & 0.00 [0.00, 0.01] & 1.00 [0.99, 1.00]\\
0.01 & 1.0 & 0.1 & 0.00 [0.00, 0.01] & 0.78 [0.75, 0.81]\\
0.01 & 1.0 & 0.2 & 0.00 [0.00, 0.00] & 0.94 [0.92, 0.95]\\
0.01 & 1.0 & 0.3 & 0.00 [0.00, 0.00] & 1.00 [0.99, 1.00]\\
0.01 & 1.0 & 0.4 & 0.00 [0.00, 0.00] & 1.00 [0.99, 1.00]\\
0.01 & 1.0 & 0.5 & 0.00 [0.00, 0.01] & 1.00 [0.99, 1.00]\\
0.01 & 1.5 & 0.1 & 0.00 [0.00, 0.01] & 0.62 [0.58, 0.65]\\
0.01 & 1.5 & 0.2 & 0.00 [0.00, 0.00] & 0.83 [0.81, 0.85]\\
0.01 & 1.5 & 0.3 & 0.00 [0.00, 0.00] & 0.92 [0.91, 0.93]\\
0.01 & 1.5 & 0.4 & 0.00 [0.00, 0.00] & 0.98 [0.97, 0.98]\\
0.01 & 1.5 & 0.5 & 0.00 [0.00, 0.01] & 1.00 [0.99, 1.00]\\
1.00 & 0.5 & 0.1 & 0.02 [0.01, 0.03] & 0.92 [0.89, 0.94]\\
1.00 & 0.5 & 0.2 & 0.04 [0.03, 0.06] & 0.82 [0.80, 0.84]\\
1.00 & 0.5 & 0.3 & 0.20 [0.19, 0.22] & 0.60 [0.58, 0.62]\\
1.00 & 0.5 & 0.4 & 0.35 [0.33, 0.38] & 0.49 [0.46, 0.51]\\
1.00 & 0.5 & 0.5 & 0.47 [0.44, 0.51] & 0.35 [0.32, 0.39]\\
1.00 & 1.0 & 0.1 & 0.01 [0.01, 0.03] & 0.73 [0.69, 0.76]\\
1.00 & 1.0 & 0.2 & 0.02 [0.01, 0.02] & 0.91 [0.89, 0.93]\\
1.00 & 1.0 & 0.3 & 0.03 [0.03, 0.04] & 0.87 [0.86, 0.89]\\
1.00 & 1.0 & 0.4 & 0.12 [0.10, 0.14] & 0.70 [0.68, 0.73]\\
1.00 & 1.0 & 0.5 & 0.27 [0.23, 0.30] & 0.54 [0.50, 0.58]\\
1.00 & 1.5 & 0.1 & 0.01 [0.01, 0.03] & 0.25 [0.21, 0.28]\\
1.00 & 1.5 & 0.2 & 0.01 [0.00, 0.01] & 0.66 [0.63, 0.69]\\
1.00 & 1.5 & 0.3 & 0.01 [0.00, 0.01] & 0.88 [0.87, 0.90]\\
1.00 & 1.5 & 0.4 & 0.03 [0.02, 0.04] & 0.88 [0.86, 0.90]\\
1.00 & 1.5 & 0.5 & 0.07 [0.05, 0.10] & 0.82 [0.79, 0.85]\\
3.00 & 0.5 & 0.1 & 0.08 [0.06, 0.11] & 0.73 [0.70, 0.77]\\
3.00 & 0.5 & 0.2 & 0.54 [0.51, 0.57] & 0.29 [0.27, 0.32]\\
3.00 & 0.5 & 0.3 & 0.85 [0.84, 0.87] & 0.02 [0.02, 0.03]\\
3.00 & 0.5 & 0.4 & 0.97 [0.96, 0.98] & 0.00 [0.00, 0.00]\\
3.00 & 0.5 & 0.5 & 1.00 [0.99, 1.00] & 0.00 [0.00, 0.01]\\
3.00 & 1.0 & 0.1 & 0.03 [0.02, 0.05] & 0.43 [0.39, 0.47]\\
3.00 & 1.0 & 0.2 & 0.11 [0.09, 0.12] & 0.70 [0.67, 0.72]\\
3.00 & 1.0 & 0.3 & 0.42 [0.40, 0.44] & 0.39 [0.36, 0.41]\\
3.00 & 1.0 & 0.4 & 0.65 [0.62, 0.68] & 0.15 [0.13, 0.17]\\
3.00 & 1.0 & 0.5 & 0.84 [0.80, 0.86] & 0.06 [0.04, 0.08]\\
3.00 & 1.5 & 0.1 & 0.02 [0.01, 0.03] & 0.09 [0.07, 0.12]\\
3.00 & 1.5 & 0.2 & 0.02 [0.01, 0.03] & 0.43 [0.41, 0.46]\\
3.00 & 1.5 & 0.3 & 0.03 [0.02, 0.03] & 0.72 [0.69, 0.74]\\
3.00 & 1.5 & 0.4 & 0.09 [0.07, 0.10] & 0.71 [0.69, 0.74]\\
3.00 & 1.5 & 0.5 & 0.25 [0.22, 0.28] & 0.52 [0.49, 0.56]\\
5.00 & 0.5 & 0.1 & 0.20 [0.17, 0.23] & 0.59 [0.55, 0.63]\\
5.00 & 0.5 & 0.2 & 0.78 [0.75, 0.80] & 0.08 [0.06, 0.09]\\
5.00 & 0.5 & 0.3 & 0.99 [0.98, 0.99] & 0.00 [0.00, 0.00]\\
5.00 & 0.5 & 0.4 & 1.00 [1.00, 1.00] & 0.00 [0.00, 0.00]\\
5.00 & 0.5 & 0.5 & 1.00 [0.99, 1.00] & 0.00 [0.00, 0.01]\\
5.00 & 1.0 & 0.1 & 0.04 [0.02, 0.05] & 0.31 [0.27, 0.34]\\
5.00 & 1.0 & 0.2 & 0.17 [0.15, 0.19] & 0.55 [0.52, 0.58]\\
5.00 & 1.0 & 0.3 & 0.65 [0.63, 0.67] & 0.16 [0.14, 0.18]\\
5.00 & 1.0 & 0.4 & 0.90 [0.88, 0.92] & 0.02 [0.01, 0.03]\\
5.00 & 1.0 & 0.5 & 0.98 [0.97, 0.99] & 0.00 [0.00, 0.01]\\
5.00 & 1.5 & 0.1 & 0.02 [0.01, 0.03] & 0.05 [0.03, 0.07]\\
5.00 & 1.5 & 0.2 & 0.02 [0.01, 0.03] & 0.33 [0.30, 0.36]\\
5.00 & 1.5 & 0.3 & 0.03 [0.03, 0.04] & 0.63 [0.61, 0.65]\\
5.00 & 1.5 & 0.4 & 0.14 [0.12, 0.16] & 0.57 [0.54, 0.60]\\
5.00 & 1.5 & 0.5 & 0.38 [0.35, 0.42] & 0.35 [0.31, 0.39]\\*
\bottomrule

\caption{\label{tbl-model3-full}Model 3 (Zipfian): fraction of seeds by
\(k\), \(\sigma\), and overlap gap, averaged over \(\mu\) and Zipf
exponent \(s\) -- the same values plotted in
Figure~\ref{fig-m3-heatmap}. Bracketed values are 95\% Wilson CIs,
pooled over all seeds contributing to each cell.}

\tabularnewline
\end{longtable}

\newpage{}

\section{Full Numerical Results}\label{sec-app-f}

Raw numerical results underlying Figure~\ref{fig-m1-heatmap} and
Figure~\ref{fig-m2-heatmap}. Values show the fraction of seeds with each
outcome, averaged over \(\mu \in \{10, 30, 60, 100\}\).

\textbf{Zero-Sensitivity Learner} (Figure~\ref{fig-m1-heatmap}). Format:
\(o_b < o_w\) / \(o_w < o_b\) / neither.

\begin{longtable}{rrlll}

\toprule
$\sigma$ & gap & $o_b < o_w$ & $o_w < o_b$ & Neither\\
\midrule
\endfirsthead
\multicolumn{5}{@{}l}{\textit{(continued)}}\\
\toprule
$\sigma$ & gap & $o_b < o_w$ & $o_w < o_b$ & Neither\\
\midrule
\endhead

\endfoot
\endlastfoot
0.5 & 0.1 & 1.00 [0.98, 1.00] & 0.00 [0.00, 0.02] & 0.00\\
0.5 & 0.2 & 1.00 [0.99, 1.00] & 0.00 [0.00, 0.01] & 0.00\\
0.5 & 0.3 & 1.00 [0.99, 1.00] & 0.00 [0.00, 0.01] & 0.00\\
0.5 & 0.4 & 1.00 [0.99, 1.00] & 0.00 [0.00, 0.01] & 0.00\\
0.5 & 0.5 & 1.00 [0.98, 1.00] & 0.00 [0.00, 0.02] & 0.00\\
1.0 & 0.1 & 0.72 [0.65, 0.78] & 0.28 [0.15, 0.26] & 0.00\\
1.0 & 0.2 & 1.00 [0.99, 1.00] & 0.00 [0.00, 0.01] & 0.00\\
1.0 & 0.3 & 1.00 [0.99, 1.00] & 0.00 [0.00, 0.01] & 0.00\\
1.0 & 0.4 & 1.00 [0.99, 1.00] & 0.00 [0.00, 0.01] & 0.00\\
1.0 & 0.5 & 1.00 [0.98, 1.00] & 0.00 [0.00, 0.02] & 0.00\\
1.5 & 0.1 & 0.03 [0.01, 0.06] & 0.52 [0.45, 0.58] & 0.44\\
1.5 & 0.2 & 0.29 [0.25, 0.34] & 0.55 [0.38, 0.47] & 0.15\\
1.5 & 0.3 & 0.68 [0.64, 0.72] & 0.32 [0.20, 0.27] & 0.00\\
1.5 & 0.4 & 0.84 [0.80, 0.87] & 0.16 [0.09, 0.16] & 0.00\\
1.5 & 0.5 & 0.95 [0.90, 0.97] & 0.05 [0.00, 0.04] & 0.00\\*
\bottomrule

\caption{\label{tbl-full-zsl}Zero-Sensitivity Learner: fraction of seeds
by \(\sigma\) and overlap gap -- the same values plotted in
Figure~\ref{fig-m1-heatmap}. Bracketed values are 95\% Wilson CIs,
pooled over all seeds contributing to each cell.}

\tabularnewline
\end{longtable}

\textbf{Variable-Sensitivity Learner} (Figure~\ref{fig-m2-heatmap}).
Format: \(o_b < o_w\) / \(o_w < o_b\) / tie / neither.

\begin{longtable}{rrrllll}

\toprule
$k$ & $\sigma$ & gap & $o_b < o_w$ & $o_w < o_b$ & Tie & Neither\\
\midrule
\endfirsthead
\multicolumn{7}{@{}l}{\textit{(continued)}}\\
\toprule
$k$ & $\sigma$ & gap & $o_b < o_w$ & $o_w < o_b$ & Tie & Neither\\
\midrule
\endhead

\endfoot
\endlastfoot
0.01 & 0.5 & 0.1 & 0.00 [0.00, 0.02] & 1.00 [0.98, 1.00] & 0.00 & 0.00\\
0.01 & 0.5 & 0.2 & 0.00 [0.00, 0.01] & 1.00 [0.99, 1.00] & 0.00 & 0.00\\
0.01 & 0.5 & 0.3 & 0.00 [0.00, 0.01] & 1.00 [0.99, 1.00] & 0.00 & 0.00\\
0.01 & 0.5 & 0.4 & 0.00 [0.00, 0.01] & 1.00 [0.99, 1.00] & 0.00 & 0.00\\
0.01 & 0.5 & 0.5 & 0.00 [0.00, 0.02] & 1.00 [0.98, 1.00] & 0.00 & 0.00\\
0.01 & 1.0 & 0.1 & 0.00 [0.00, 0.02] & 0.99 [0.97, 1.00] & 0.00 & 0.00\\
0.01 & 1.0 & 0.2 & 0.00 [0.00, 0.01] & 1.00 [0.99, 1.00] & 0.00 & 0.00\\
0.01 & 1.0 & 0.3 & 0.00 [0.00, 0.01] & 1.00 [0.99, 1.00] & 0.00 & 0.00\\
0.01 & 1.0 & 0.4 & 0.00 [0.00, 0.01] & 1.00 [0.99, 1.00] & 0.00 & 0.00\\
0.01 & 1.0 & 0.5 & 0.00 [0.00, 0.02] & 1.00 [0.98, 1.00] & 0.00 & 0.00\\
0.01 & 1.5 & 0.1 & 0.00 [0.00, 0.02] & 0.76 [0.70, 0.82] & 0.00 & 0.24\\
0.01 & 1.5 & 0.2 & 0.00 [0.00, 0.01] & 0.97 [0.95, 0.98] & 0.00 & 0.03\\
0.01 & 1.5 & 0.3 & 0.00 [0.00, 0.01] & 1.00 [0.99, 1.00] & 0.00 & 0.00\\
0.01 & 1.5 & 0.4 & 0.00 [0.00, 0.01] & 1.00 [0.99, 1.00] & 0.00 & 0.00\\
0.01 & 1.5 & 0.5 & 0.00 [0.00, 0.02] & 1.00 [0.98, 1.00] & 0.00 & 0.00\\
0.10 & 0.5 & 0.1 & 0.00 [0.00, 0.02] & 1.00 [0.98, 1.00] & 0.00 & 0.00\\
0.10 & 0.5 & 0.2 & 0.00 [0.00, 0.01] & 1.00 [0.99, 1.00] & 0.00 & 0.00\\
0.10 & 0.5 & 0.3 & 0.00 [0.00, 0.01] & 1.00 [0.99, 1.00] & 0.00 & 0.00\\
0.10 & 0.5 & 0.4 & 0.00 [0.00, 0.01] & 0.99 [0.97, 0.99] & 0.01 & 0.00\\
0.10 & 0.5 & 0.5 & 0.00 [0.00, 0.02] & 0.98 [0.96, 0.99] & 0.02 & 0.00\\
0.10 & 1.0 & 0.1 & 0.00 [0.00, 0.02] & 0.99 [0.97, 1.00] & 0.00 & 0.00\\
0.10 & 1.0 & 0.2 & 0.00 [0.00, 0.01] & 1.00 [0.99, 1.00] & 0.00 & 0.00\\
0.10 & 1.0 & 0.3 & 0.00 [0.00, 0.01] & 1.00 [0.99, 1.00] & 0.00 & 0.00\\
0.10 & 1.0 & 0.4 & 0.00 [0.00, 0.01] & 1.00 [0.99, 1.00] & 0.00 & 0.00\\
0.10 & 1.0 & 0.5 & 0.00 [0.00, 0.02] & 0.99 [0.97, 1.00] & 0.00 & 0.00\\
0.10 & 1.5 & 0.1 & 0.00 [0.00, 0.02] & 0.72 [0.65, 0.78] & 0.00 & 0.28\\
0.10 & 1.5 & 0.2 & 0.00 [0.00, 0.01] & 0.94 [0.92, 0.96] & 0.00 & 0.06\\
0.10 & 1.5 & 0.3 & 0.00 [0.00, 0.01] & 0.99 [0.99, 1.00] & 0.00 & 0.00\\
0.10 & 1.5 & 0.4 & 0.00 [0.00, 0.01] & 1.00 [0.99, 1.00] & 0.00 & 0.00\\
0.10 & 1.5 & 0.5 & 0.00 [0.00, 0.02] & 1.00 [0.98, 1.00] & 0.00 & 0.00\\
0.50 & 0.5 & 0.1 & 0.00 [0.00, 0.03] & 0.94 [0.90, 0.97] & 0.05 & 0.00\\
0.50 & 0.5 & 0.2 & 0.33 [0.28, 0.37] & 0.33 [0.28, 0.37] & 0.34 & 0.00\\
0.50 & 0.5 & 0.3 & 0.67 [0.63, 0.71] & 0.23 [0.20, 0.27] & 0.10 & 0.00\\
0.50 & 0.5 & 0.4 & 0.76 [0.71, 0.80] & 0.14 [0.11, 0.18] & 0.10 & 0.00\\
0.50 & 0.5 & 0.5 & 0.77 [0.71, 0.82] & 0.05 [0.03, 0.09] & 0.18 & 0.00\\
0.50 & 1.0 & 0.1 & 0.02 [0.01, 0.05] & 0.92 [0.87, 0.95] & 0.06 & 0.00\\
0.50 & 1.0 & 0.2 & 0.15 [0.12, 0.19] & 0.55 [0.50, 0.60] & 0.30 & 0.00\\
0.50 & 1.0 & 0.3 & 0.52 [0.48, 0.56] & 0.26 [0.23, 0.30] & 0.22 & 0.00\\
0.50 & 1.0 & 0.4 & 0.72 [0.67, 0.76] & 0.21 [0.17, 0.25] & 0.07 & 0.00\\
0.50 & 1.0 & 0.5 & 0.75 [0.69, 0.80] & 0.15 [0.11, 0.21] & 0.10 & 0.00\\
0.50 & 1.5 & 0.1 & 0.01 [0.00, 0.04] & 0.60 [0.54, 0.67] & 0.00 & 0.38\\
0.50 & 1.5 & 0.2 & 0.03 [0.02, 0.05] & 0.78 [0.74, 0.82] & 0.07 & 0.12\\
0.50 & 1.5 & 0.3 & 0.16 [0.13, 0.19] & 0.62 [0.58, 0.66] & 0.22 & 0.00\\
0.50 & 1.5 & 0.4 & 0.38 [0.33, 0.43] & 0.35 [0.30, 0.40] & 0.27 & 0.00\\
0.50 & 1.5 & 0.5 & 0.56 [0.50, 0.63] & 0.26 [0.20, 0.32] & 0.18 & 0.00\\
1.00 & 0.5 & 0.1 & 0.53 [0.46, 0.60] & 0.21 [0.16, 0.28] & 0.26 & 0.00\\
1.00 & 0.5 & 0.2 & 0.89 [0.86, 0.92] & 0.00 [0.00, 0.01] & 0.11 & 0.00\\
1.00 & 0.5 & 0.3 & 1.00 [0.99, 1.00] & 0.00 [0.00, 0.01] & 0.00 & 0.00\\
1.00 & 0.5 & 0.4 & 1.00 [0.99, 1.00] & 0.00 [0.00, 0.01] & 0.00 & 0.00\\
1.00 & 0.5 & 0.5 & 1.00 [0.98, 1.00] & 0.00 [0.00, 0.02] & 0.00 & 0.00\\
1.00 & 1.0 & 0.1 & 0.08 [0.05, 0.13] & 0.74 [0.68, 0.80] & 0.18 & 0.00\\
1.00 & 1.0 & 0.2 & 0.63 [0.58, 0.68] & 0.16 [0.13, 0.20] & 0.21 & 0.00\\
1.00 & 1.0 & 0.3 & 0.90 [0.88, 0.93] & 0.01 [0.00, 0.02] & 0.09 & 0.00\\
1.00 & 1.0 & 0.4 & 1.00 [0.99, 1.00] & 0.00 [0.00, 0.01] & 0.00 & 0.00\\
1.00 & 1.0 & 0.5 & 1.00 [0.98, 1.00] & 0.00 [0.00, 0.02] & 0.00 & 0.00\\
1.00 & 1.5 & 0.1 & 0.02 [0.01, 0.05] & 0.52 [0.46, 0.59] & 0.02 & 0.43\\
1.00 & 1.5 & 0.2 & 0.13 [0.10, 0.17] & 0.58 [0.53, 0.63] & 0.14 & 0.14\\
1.00 & 1.5 & 0.3 & 0.42 [0.38, 0.46] & 0.34 [0.30, 0.37] & 0.24 & 0.01\\
1.00 & 1.5 & 0.4 & 0.66 [0.61, 0.70] & 0.22 [0.18, 0.26] & 0.12 & 0.00\\
1.00 & 1.5 & 0.5 & 0.76 [0.70, 0.82] & 0.15 [0.11, 0.21] & 0.08 & 0.00\\*
\bottomrule

\caption{\label{tbl-full-vsl}Variable-Sensitivity Learner: fraction of
seeds by \(k\), \(\sigma\), and overlap gap -- the same values plotted
in Figure~\ref{fig-m2-heatmap}. Bracketed values are 95\% Wilson CIs,
pooled over all seeds contributing to each cell.}

\tabularnewline
\end{longtable}

\newpage{}

\section{Threshold Robustness}\label{sec-app-g}

The \(o_b\) criterion depends on two thresholds: CLASS\_FRAC (the
minimum fraction of verbs that must pass the Mann-Whitney test) and the
Mann-Whitney significance cutoff \(p < 0.001\). The \(o_w\) criterion
depends on DJS\_THRESH (the absolute within-class \(D_\text{JS}\)
threshold) and SUSTAIN (the required number of consecutive steps above
threshold, fixed at 3 throughout). Baseline values are CLASS\_FRAC
\(= 0.10\) and DJS\_THRESH \(= 0.01\).

We note that these thresholds are not directly taken from Jian and
Manning (2026)'s Experiment 1: J\&M identify onsets visually from
plotted trajectories for GPT-2 and do not specify a formal CLASS\_FRAC
for \(o_b\) or a formal DJS\_THRESH for \(o_w\) in that experiment. The
DJS\_THRESH value is borrowed from their Experiment 2 breakpoint
criterion (sustained \(D_\text{JS}\) at least 0.01 above a per-verb
baseline), applied here to mean within-class \(D_\text{JS}\) across verb
pairs. A direct calibration check against J\&M's GPT-2 trajectories is
not possible: their intermediate training checkpoints are not publicly
available.

To assess robustness, we precomputed the full \(D_\text{JS}\)-matrix
trajectory (50 seeds each) for two representative parameter
combinations: one that produces \(o_b < o_w\) (Zero-Sensitivity Learner,
\(\mu=100\), \(\sigma=0.5\), item\_overlap \(= 0.7\), class\_overlap
\(= 0.2\)) and one that produces \(o_w < o_b\) (Variable-Sensitivity
Learner, \(k=0.001\), same structural parameters). We then applied onset
detection across a grid of CLASS\_FRAC
\(\in \{0.01, 0.05, 0.10, 0.20, 0.30\}\) and DJS\_THRESH
\(\in \{0.001, 0.005, 0.010, 0.020, 0.050\}\) without recomputing the
DJS matrices.

\begin{table}[H]

\centering{

\begin{tabular}{lrrrrr}
\toprule
  & DJS = 0.001 & DJS = 0.005 & DJS = 0.01 & DJS = 0.02 & DJS = 0.05\\
\midrule
CLASS\_FRAC = 0.01 & 1.00 & 1.00 & 1.00 & 1.00 & 1.00\\
CLASS\_FRAC = 0.05 & 1.00 & 1.00 & 1.00 & 1.00 & 1.00\\
CLASS\_FRAC = 0.1 & 1.00 & 1.00 & 1.00 & 1.00 & 1.00\\
CLASS\_FRAC = 0.2 & 1.00 & 1.00 & 1.00 & 1.00 & 1.00\\
CLASS\_FRAC = 0.3 & 1.00 & 1.00 & 1.00 & 1.00 & 1.00\\
\bottomrule
\end{tabular}

}

\caption{\label{tbl-robustness-case1}Case 1 (Zero-Sensitivity Learner,
\(o_b < o_w\)): fraction of seeds with \(o_b < o_w\) across all
threshold combinations. \(o_w < o_b = 0.00\) and neither \(= 0.00\)
throughout.}

\end{table}%

\begin{table}[H]

\centering{

\begin{tabular}{lrrrrr}
\toprule
  & DJS = 0.001 & DJS = 0.005 & DJS = 0.01 & DJS = 0.02 & DJS = 0.05\\
\midrule
CLASS\_FRAC = 0.01 & 1.00 & 1.00 & 1.00 & 1.00 & 1.00\\
CLASS\_FRAC = 0.05 & 1.00 & 1.00 & 1.00 & 1.00 & 1.00\\
CLASS\_FRAC = 0.1 & 1.00 & 1.00 & 1.00 & 1.00 & 1.00\\
CLASS\_FRAC = 0.2 & 1.00 & 1.00 & 1.00 & 1.00 & 1.00\\
CLASS\_FRAC = 0.3 & 1.00 & 1.00 & 1.00 & 1.00 & 1.00\\
\bottomrule
\end{tabular}

}

\caption{\label{tbl-robustness-case2}Case 2 (Variable-Sensitivity
Learner, \(k = 0.001\), \(o_w < o_b\)): fraction of seeds with
\(o_w < o_b\), which holds in 100\% of seeds across every threshold
combination tested; \(o_b < o_w = 0.00\) throughout. Results are
invariant to both CLASS\_FRAC and DJS\_THRESH because within-class
\(D_{\text{JS}}\) spikes to \(\approx 0.5\) within the first few
observations, trivially exceeding any tested DJS\_THRESH, while \(o_b\)
requires several observations to accumulate even a single passing verb
at any tested CLASS\_FRAC.}

\end{table}%

Both orderings are perfectly robust to threshold choice: \(o_b < o_w\)
holds in 100\% of seeds across all 25 combinations in Case 1, and
\(o_w < o_b\) holds in 100\% of seeds across all 25 combinations in Case
2; the reverse ordering never occurs in either case, including at our
baseline CLASS\_FRAC \(= 0.10\). The results are invariant to
DJS\_THRESH because at low \(k\), within-class \(D_\text{JS}\)
immediately far exceeds any reasonable threshold, and invariant to
CLASS\_FRAC because between-class structure takes measurably longer to
accumulate regardless of how permissive the threshold is set.

\bibliography{references.bib}

\end{document}